\documentclass{ecai}
\usepackage{graphicx}
\usepackage{latexsym}

\usepackage{amsmath}
\usepackage{booktabs}
\usepackage{makecell}
\usepackage{tikz}
\usepackage{amsfonts}
\ecaisubmission   

\begin{document}

\begin{frontmatter}

\title{GenCo: An Auxiliary Generator from Contrastive Learning for Enhanced Few-Shot Learning in Remote Sensing}

\author[A]{~\snm{Jing Wu}\orcid{0000-0001-8619-566X}\thanks{Corresponding Author. Email: {jingwu6}@illinois.edu.}}
\author[A]{~\snm{Naira Hovakimyan}}
\author[B]{~\snm{Jennifer Hobbs}} 

\address[A]{University of Illinois Urbana-Champaign}
\address[B]{Intelinair}


\begin{abstract}
Classifying and segmenting patterns from a limited number of examples is a significant challenge in remote sensing and earth observation due to the difficulty in acquiring accurately labeled data in large quantities. Previous studies have shown that meta-learning, which involves episodic training on query and support sets, is a promising approach. However, there has been little attention paid to direct fine-tuning techniques. This paper repurposes contrastive learning as a pre-training method for few-shot learning for classification and semantic segmentation tasks. Specifically, we introduce a generator-based contrastive learning framework (GenCo) that pre-trains backbones and simultaneously explores variants of feature samples. In fine-tuning, the auxiliary generator can be used to enrich limited labeled data samples in feature space. We demonstrate the effectiveness of our method in improving few-shot learning performance on two key remote sensing datasets: Agriculture-Vision and EuroSAT. Empirically, our approach outperforms purely supervised training on the nearly 95,000 images in Agriculture-Vision for both classification and semantic segmentation tasks. Similarly, the proposed few-shot method achieves better results on the land-cover classification task on EuroSAT compared to the results obtained from fully supervised model training on the dataset.
\end{abstract}

\end{frontmatter}

\section{Introduction}
%

Remote sensing (RS) and earth observation (EO) imagery enable the detection and monitoring of critical societal challenges, including food security, natural disasters, clean water availability, hunger and poverty, the impact of climate change, threats to animal habitats, geopolitical risk, and more.  Like other domains, RS-EO has benefited from significant advances in machine vision systems over the past decade.  However, these algorithms' ability to learn without abundant labeled data is far from satisfactory, even when compared to a toddler. This fact severely limits the scalability of learned models, with various categories following the long-tail distribution in the real world. The lack of labeled samples is even more severe for RS-EO tasks, as data acquisition often involves concerns around security, ethics, resources, accessibility, and cost \cite{yang2022survey}. Furthermore, annotation of remote sensing data often requires high levels of expert knowledge and true ground-truth verification (i.e. physically traveling to locations to confirm predictions). Therefore developing machine-learning applications from RS-EO data to address key societal challenges is often thwarted by the domain's particularly laborious dataset-creation process~\cite{sun2021research,street2sat}.

Inspired by human's highly efficient learning ability, research around learning from unlabeled data, i.e. unsupervised or self-supervised learning, and the ability to generalize from only a few examples, i.e. few-shot learning, have become key areas of interest in the machine learning community~\cite{vinyals2016matching,finn2017model,sung2018learning,gidaris2018dynamic,sun2021research,alajaji2020few,li2021few}. Few-shot learning aims to realize the knowledge adaption of embeddings from label-abundant data to label-scarce classes. While the adapted representation aims to discriminate different levels of information (e.g., instance level and semantic level) between classes, the embedding should be invariant to common, irrelevant variations of the image, including different sizes, deformations, and lighting. The question is then: How can we learn a representation invariant to common factors while maintaining differences for diverse classes with limited labels?

As a prevailing and advancing research topic, contrastive learning has demonstrated impressive results on various downstream learning tasks. These methods seek to learn a transferable representation by strongly augmenting large quantities of raw data and pulling views of the same image close together while pushing differing images apart. These methods are often evaluated on the performance of downstream tasks fine-tuned on different fractions of the supervised dataset. However, the focus is rarely on the one-shot or few-shot cases for extremely limited datasets. As raw RS-EO data is highly abundant, but ground-truth data is extremely scarce, leveraging contrastive methods for few-shot learning offers a key opportunity in this domain. Additionally, most common contrastive learning and few-shot methods were developed for natural scene imagery; e.g. \cite{he2020momentum,chen2020simple,grill2020bootstrap,zbontar2021barlow} show that as the statistics of that domain (both source imagery and targets) are extremely different from RS-EO data, there is no guarantee that the same benefit will be observed without adaptation. Therefore, we investigate the improvement in the performance of few-shot learning in RS-EO classification and semantic segmentation tasks using contrastive-learning-based pre-training.

Specifically, we focus on pre-training from the Extended Agriculture Vision dataset (AV+) \cite{wu2023extended}, which includes high-resolution aerial imagery over agricultural lands in the US Midwest. Obtaining ground-truth annotations for agriculture is particularly challenging due to patterns of interest being small in size, high in number, and often possessing ambiguous boundaries; the ability to identify patterns from only a small number of samples addresses key challenges in precision agriculture and food security.

Drawing inspiration from the work of \cite{gidaris2018dynamic, qi2018low, wang2020frustratingly}, we have adopted a two-stage training approach that involves contrastive-learning-based pre-training followed by fine-tuning. Specifically, we have developed a contrastive learning framework, GenCo, with an auxiliary generator trained. During the pre-training, the generator is tasked with exploring variants of encoded features and formulating additional positive pairs. During the fine-tuning stage, we fixed the parameters of the encoders and trained only the classification layer and decoder for classification and segmentation, respectively. Notably, only a limited number of labeled data samples are provided during the fine-tuning stage. To address this, we introduced the generator from the contrastive learning model to create further samples in feature space and enrich knowledge during the few-shot tasks.

We find that the embeddings pre-trained from AV+ under this protocol show better adaptability when compared to counterparts pre-trained on ImageNet~\cite{deng2009imagenet} and COCO~\cite{lin2014microsoft}. Our method outperforms pre-trained ImageNet weights by 1 to 6 points on Agriculture-Vision and EuroSAT classification tasks under the same number of supervised training samples. Similarly, our embeddings deliver a 5 to 7-point improvement on mIoU compared with embeddings learned from COCO on the Agriculture-Vision semantic segmentation task. 

Meanwhile, we demonstrate the high learning efficiency of the proposed method for RS-EO imagery. With a few labeled images, we find that the GenCo shows comparable or even better results under different tasks and datasets when compared with supervised models trained on fully labeled data samples; our proposed approach shows matching performance with less than 0.01 percent of labeled data.

In summary, the contributions of this paper can be summarized:

(1) We leverage contrastive learning for both classification and segmentation of few-shot learning tasks using remote sensing imagery and demonstrate the successful adaptation of the two-stage contrastive-learning-based few-shot strategy for RS-EO data.

(2) We propose a novel generative-based contrastive learning framework GenCo that benefits both pre-training and the downstream few-shot classification and segmentation. With negligible computation, the designed generator enhances the quality of embeddings and simultaneously provides extra data samples for downstream few-shot tasks.

(3) Extensive experiments show that our approach allows us to competitively identify key agricultural and land cover patterns with only a small amount of labeled data.

(4) We demonstrate that pre-training on AV+, a high-resolution multi-spectral RS-EO dataset, provides strong benefits to other RS-EO tasks on lower-resolution data such as EuroSAT.

\section{Related Work}

\subsection{Few-Shot Learning}
Among various methods to speed up training and enhance label efficiency, such as neural network pruning\cite{luo2017thinet,wang2023double}, simulation-based training\cite{wu2022optimizing,tao2022optimizing}, or few-shot learning. Few-shot learning is the most promising one in various application domains like healthcare\cite{chen2021momentum,chen2019claims}, remote sensing\cite{sun2021research}, for learning tasks including classification \cite{fei2006one,wang2022global}   object detection \cite{wang2020frustratingly}, semantic segmentation \cite{wang2019panet}, and robot learning \cite{finn2017one}. Generally, previous works can be roughly cast into three categories: metric-based, optimization-based, and hallucination-based. 

The key idea of metric-based approaches is to learn good embeddings with appropriate kernels. Previous results from \cite{koch2015siamese} propose applying a siamese neural network for few-shot classification. Following that, \cite{sung2018learning} presents a Relation Network by replacing the L1 distance between features with a convolutional neural network (CNN)-based classifier and updating the mean squared error (MSE) with cross-entropy; the  triplet loss is utilized to improve the model's performance \cite{cacheux2019modeling}. \cite{gidaris2019generating} further adds extra self-supervised tasks to enhance generalization capacity.

Optimization-based methods aim to learn through gradient backpropagation. Representative works include MAML \cite{finn2017model}, which realizes quick adaption from good initialization,  Reptile \cite{nichol2018reptile}, which simplifies the learning process of MAML, and MetaOptNet \cite{lee2019meta}, which incorporates the support vector machine (SVM) as a classifier. 

Hallucination-based methods seek to learn generators to generate unseen samples. Works from \cite{wang2018low,zhang2019few,li2020adversarial} show that such a strategy of hallucination improves the test results and enhances the generation of models.

Few-Shot learning for RS-EO has received increased attention in recent years,~\cite{sun2021research}. While much of the work is focused on scene classification~\cite{alajaji2020few,li2020dla,alajaji2020meta,liu2018deep}, other recent approaches examine semantic segmentation tasks ~\cite{wang2021dmml,kemker2018low,yao2021scale}.

\begin{figure*}[t!]
\begin{center}
 \includegraphics[width=0.8\textwidth]{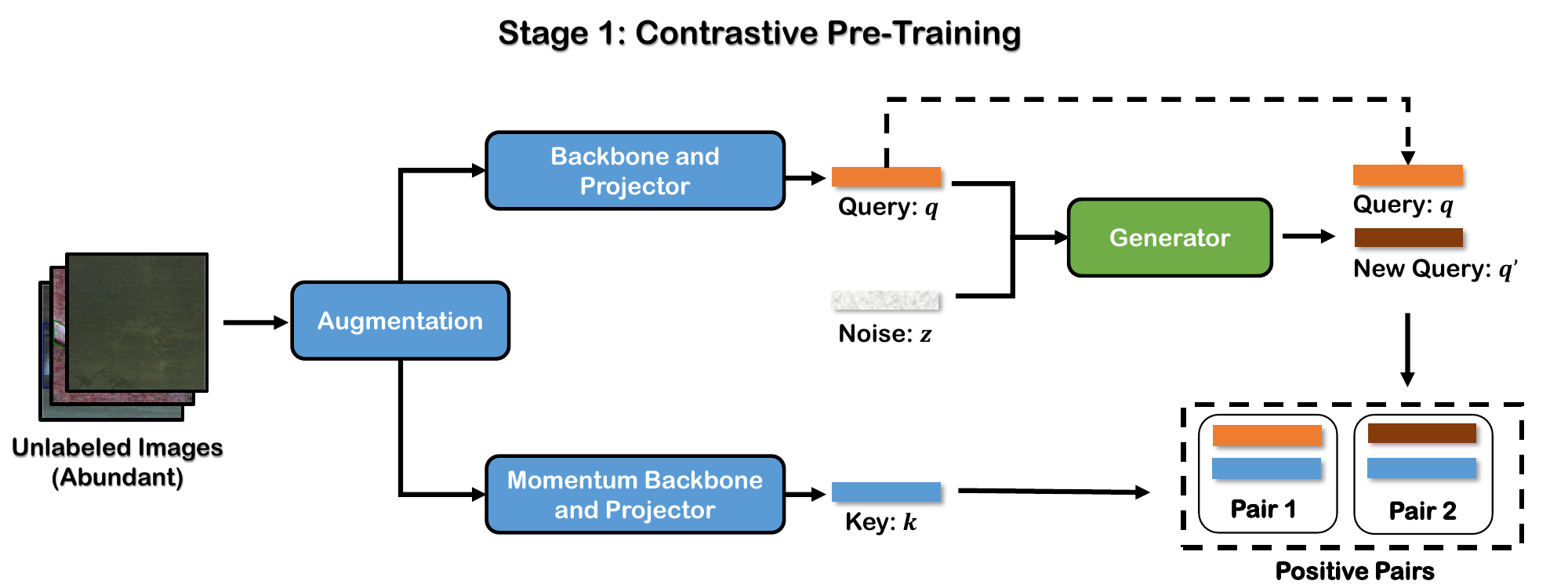}
\end{center}
\vspace*{-5mm}
  \caption{Illustration of the pre-training stage of GenCo. In the pre-training stage, we follow the training strategy of contrastive learning to train the backbones and projectors jointly with abundant unlabeled images. An additional feature vector $q'$ is added by passing the real feature vector $q$ and a noise vector $z$ to the generator. The generator is trained end-to-end along with the contrastive learning framework}
\label{fig: pre-training}
\end{figure*}

\subsection{Contrastive Learning}
Contrastive learning is widely used for pre-training without labeled data and shows superior performance in various self-supervised learning (SSL) tasks~\cite{chen2020improved,grill2020bootstrap,chen2020simple,chen2020big}. Specifically, contrastive approaches train architectures by bringing the representation of different views of the same image closer together while spreading representations of views from different images apart. The success of training may rely on large batch sizes~\cite{chen2020simple}, a momentum-update memory bank~\cite{he2020momentum}, projection heads~\cite{chen2020improved}, and/or stop-gradient trick~\cite{chen2021exploring}. Barlow Twins ~\cite{zbontar2021barlow} further proposes a new contrastive learning objective without using the trick of stop-gradient, which also brings equivalent results. Among all these methods, MoCo-V2 is one of the most widely used frameworks, given its memory efficiency and promising performance\cite{chen2020improved}. Within the field of RS-EO specifically, \cite{manas2021seasonal} takes MoCo-V2 as the basis and uses multiple projection heads to capture the desired invariance to seasonality. Therefore, we continue to utilize MoCo-V2 as a pre-training protocol in this work.

\subsection{Data Augmentation}
Data augmentation is essential for both contrastive learning and few-shot learning. Contrastive learning seeks to learn invariances in the representations to multiple augmentations so that these representations can be generalized successfully to various downstream tasks; to learn embeddings that are invariant to spatial and lighting transformations, the training of contrastive learning requires strong data augmentations, including color jittering, random flips, grayscale conversion, rotation, cropping, and resizing~\cite{chen2020improved,he2020momentum,grill2020bootstrap,chen2020simple}. Similar data augmentations are used in few-shot learning but with additional techniques like image hallucination~\cite{wang2018low}, manifold mixup~\cite{verma2019manifold}, and gallery image pool sampling~\cite{chen2019image}. More recent and related research to our paper comes from\cite{wu2023hallucination}, which demonstrated effectiveness of feature-level operations in representation learning.

\subsection{Pairing Supervised and Self-Supervised Learning}
Some methods utilize the SSL loss as  supplemental losses during the supervised training process~\cite{gidaris2019boosting,su2020does}. Often, additional efforts are needed to calibrate (i.e. re-weight) these losses when crossing different domains. More straightforward and effective methods come from supervised fine-tuning~\cite{doersch2020crosstransformers}. While SSL encourages the learning of general-purpose features, the adaption of features on the new task can be realized with only a few labeled samples.  Within RS-EO, very recent work has looked to combine SSL and few-shot learning for scene classification~\cite{zeng2022task} and segmentation~\cite{li2022global}.

\section{Method}
In this section, we first introduce the overall pipeline of pre-training and downstream few-shot learning in Section~\ref{sec: pipline}. Following this, we describe the proposed GenCo in Section~\ref{sec: pre-training}. Lastly, we demonstrate how the generator from the contrastive learning model helps the downstream few-shot task by generating extra samples in feature space ~\ref{sec: downstream}.

\subsection{The Pipeline of Learning}
\label{sec: pipline}

In the first stage, we pre-trained different backbones with contrastive learning models on unlabeled data, as shown in Figure~\ref{fig: pre-training}. To be more specific, there are roughly 1,300,000 images with shapes 512 $\times$ 512. These images are all randomly cropped from the raw images from AV+. We use all these unlabeled images as input to pre-train backbones without any supervision. After the pertaining, we move to the second stage, which fine-tunes the pre-trained backbones for the downstream few-shot tasks. Since the proposed contrastive learning is unsupervised, there is no information on any base classes, unlike the usual settings for few-shot learning. Only k samples are provided for fine-tuning during the evaluation, where k varies from 1-10. The evaluation will include both base and novel classes based on different downstream tasks to optimize the classification accuracy or mIoU of agricultural patterns on Agriculture-Vision and land covers on EuroSAT. 

\subsection{Pre-training with Generator-based Contrastive Learning}
\label{sec: pre-training}

\textbf{Basic Framework.} 
The framework of GenCo is shown in Figure~\ref{fig: pre-training}. Specifically, GenCo can be trained with natural scene images that contain information about red, green, and blue channels similar to previous works\cite{chen2020simple,chen2020improved,chen2021exploring}. However, AV+  has extra information in the NIR channel. To fully explore knowledge from the pre-training dataset, we further add one channel to the backbones following the work of from\cite{wu2023extended}.  

In every training iteration, a training sample $x$ is augmented into two different views named query $x^{q}$ and key $x^{k}$. These views contain the same semantic meaning but also variations introduced from data augmentations, including spatial and color transforms. With an online network and a momentum-updated offline network proposed in \cite{he2020momentum}, the training encourages these two views to be mapped into two similar embedding spaces, i.e., $q$, $k$, as a positive pair. For feature vectors that are not encoded from $x$, we define them as $k^{-}$  

\textbf{Feature Generation.}
Sufficient positive pairs in feature space help the performance of contrastive learning\cite{chen2020simple}. However, the large number of positive features relies on large batch sizes that may not always be accessible. To address this, we design a generator $G$ that takes the query feature $q$ and random noise $z$ as input. Entries of $z$ are assumed to follow a normal distribution with mean 0 and variance 0.1. We then generate a new sample $q'$ in the feature space where $q' = G(q,z|\theta)$. Notably, $G$ is a lightweight module with parameters $\theta$. It is instantiated with three linear layers and a ReLU layer between two successive layers.

With this simple design, we eventually obtain an additional positive pair $p(q',k)$. Along with the original positive $p(q,k)$ one, the framework provided enhanced contrast and improve the qualify of embeddings during the training with an additional little computation. Together, based on positive and negative pairs and a temperature parameter $\tau$ for scaling, the training loss function, i.e., InfoNCE \cite{oord2018representation}, is then defined as follows: 

\begin{align}
\setlength\abovedisplayskip{0pt}
\setlength\belowdisplayskip{0pt}
    \label{eq:InfoNEC}
    \mathcal{L} = -\log \frac{\exp(q\cdot k / \tau) + \exp(q'\cdot k / \tau)}{\sum_{k^-}\exp(q\cdot k^- / \tau)+\exp(q\cdot k / \tau)+ \exp(q'\cdot k / \tau)}
\end{align}

\begin{figure*}[t!]
\begin{center}
 \includegraphics[width=0.8\textwidth]{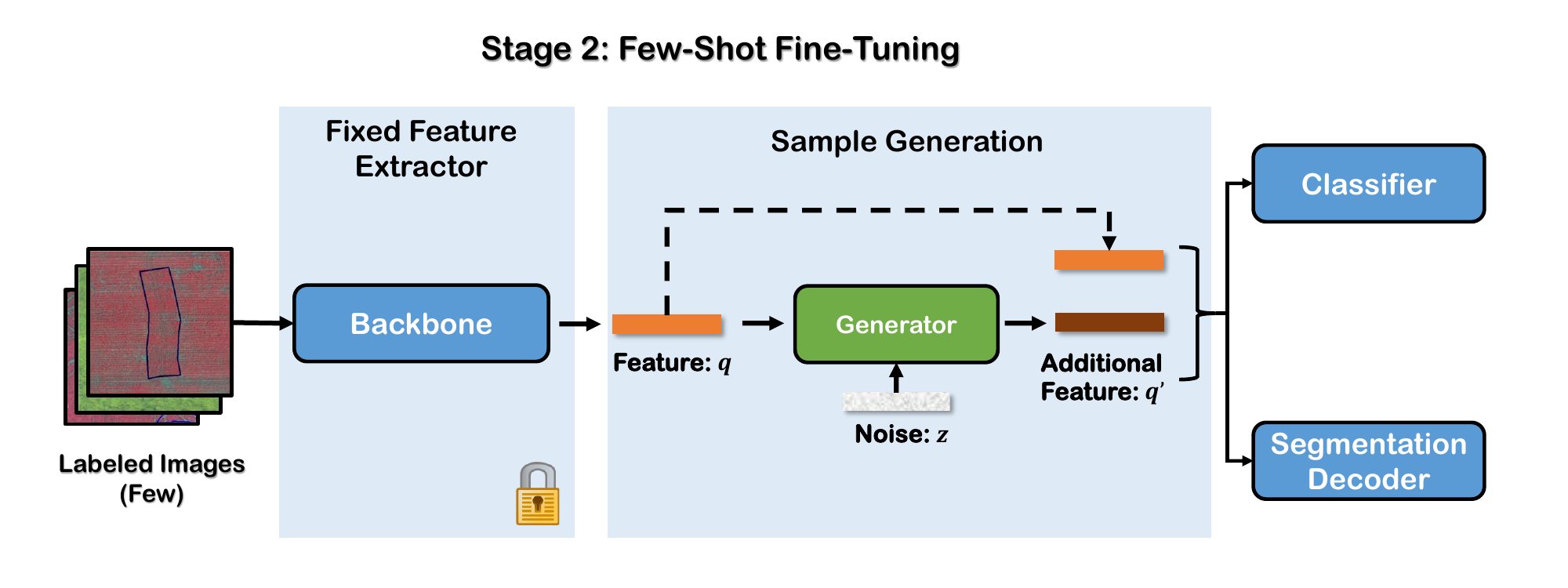}
\end{center}
\vspace*{-5mm}
  \caption{Illustration of the fine-tuning stage. In the fine-tuning stage with limited labeled images, we adapt the pre-trained backbone from stage 1 for feature extraction. The generator is brought from the GenCo to generate additional labeled data. During the tuning, the feature extractors are fixed, and fine-tuning is only involved in the generator, classifier and segmentation decoder.}
\label{fig: fine-tune}
\end{figure*}

\subsection{Downstream Few-Shot Learning with Generator}
\label{sec: downstream}
Given the strong data augmentations in contrastive learning, the backbone features encoded can capture the key representation features. It should adapt to different data classes and types of downstream tasks without much learning effort. Therefore, one of the critical steps of the proposed method is to separate representation learning and downstream task learning into two stages.

With contrastive learning applied in the first stage, we freeze the encoder and fine-tune models with a few labeled images. We first create a small balanced training set with K images per class, i.e., K shots. These classes can either be seen or be novel classes. Then, each image was encoded into feature space, forming a vector $q$ with label $y$. While the traditional fine-tuning method applied the feature $q$ directly to downstream tasks, we take advantage of the generator $G$ from the contrastive learning framework to generate labeled samples following the equation $q' = G(q,z|\theta)$ as shown in Figure~\ref{fig: fine-tune}. The $q'$ share the same label $y$ with $q$, forming an extra label data $(q', y)$. In other words, we eventually obtained 2K labeled data points given K images for each class, largely enriching the information for downstream learning tasks. During the fine-tuning stage, the generator $G$ is differentiable and optimized simultaneously with the fine-tuning stage.

In the few-shot classification task, we add one fully connected layer to the backbones without introducing extra non-linearity.  We assign randomly initialized weights to the added classification layer. In the few-shot semantic segmentation task, we choose the lightweight segmentation model U-Net \cite{ronneberger2015u} for fine-tuning given limited training samples. Concretely, we add a five-layer decoder based on the encoders pre-trained from contrastive learning. In each layer of the decoder, it performs up-sampling by a 2$\times$2 deconvolution layer to recover the original images' sizes. While the size of the feature images increases, the number of channels reduces by half after each up-sampling. Empirically, this asymmetric encoder-decoder shows exceptional performance in the segmentation task on Agriculture-Vision.

\section{Experiments and Results}
In this study, we conduct various experiments to prove the effectiveness of our proposed methods.  In Section ~\ref {sec: datasets}, we introduced the used datasets in this paper. Then, we illustrate the evaluation metrics in Section ~\ref{sub:Preliminaries}. In the following Section~\ref {sec: imp}, we demonstrate the necessary details to reproduce the experiments. Lastly, in Section~\ref{sec: results}, we report all the results based on the proposed metrics.

\subsection{Datasets}
\label{sec: datasets}
\subsubsection{Agriculture-Vision}
Agriculture-Vision (AV) is a large aerial image database for agricultural pattern analysis. It contains 94,986 high-quality images over 3432 farmlands across the US. Totally, there are nine classes selected in the dataset under the advisement of agronomists, which include double plant, dry-down, endrow, nutrient deficiency, planter skip, storm damage, water, waterway, and weed cluster. With extreme label imbalance across categories, it is a challenge to train well-performing models for classification and segmentation tasks ~\cite{Chiu_2020_CVPR_Workshops}.  The original dataset is designed for semantic segmentation; we also create a "classification" version of the dataset by assigning a positive label if any presence of that class is included in the tile.

\subsubsection{Extended Agriculture-Vision}
For contrastive pre-training, we use the large-scale remote-sensing dataset, Extend Agriculture-Vision \cite{wu2023extended}. While the original Agriculture-Vision dataset contained only 512x512 tiles with semantic segmentation labels for agricultural patterns such as waterways, weeds, nutrient deficiency, etc., AV+ includes several thousand additional raw full-field images (upwards of 10,000 x 10,000 in dimension).
Images consist of RGB and Near-infrared (NIR) channels with resolutions as high as 10 cm per pixel. As it also covers data that varies from 2017 to 2020, encoders pre-trained on this dataset should capture remote sensing, agriculture, and temporal features. Therefore, the embeddings pre-trained on Extend Agriculture-Vision should be adapted well to diverse downstream tasks such as agricultural pattern recognition and land-cover classification. 

\subsubsection{EuroSAT}
EuroSAT is a dataset for the classification task of land use and land cover. All the satellite images are collected from Sentinel-2, covering 34 countries. There are 27,000 images in total, with ten types of labels corresponding to different land use cases. The class labels are evenly distributed, with each category consisting of 2,000 to 3,000 images. We use the split method for training and evaluation following the work from \cite{manas2021seasonal,helber2019eurosat,wu2023extended}. While there is a total of 13 channels in total, in this work, we focus on the RGB channels since it is a more general modality.

\subsection{Experiment Setup and Evaluation Metrics}
\label{sub:Preliminaries}
We evaluate the proposed method from two perspectives, i.e., the quality of embeddings and the required amount of labeled data for model adaptation. First, in the few-shot classification task, we compare the performance of features from the backbone pre-trained on ImageNet, and the backbone pre-trained on AV+ using contrastive learning models. Similarly, in the few-shot semantic segmentation task, our embeddings pre-trained on AV+ and embeddings from COCO's weights are compared\cite{lin2014microsoft}. Second, to illustrate the learning efficiency of our method, we compare the two models' performances, one of which is fine-tuned on a few samples, and the other is trained in a supervised way with complete labeled data.

\subsection{Implementations Details}
\label{sec: imp}
Firstly, we introduce implementation details of the generator-based contrastive learning model to pre-train backbones, as shown in the left part of Figure\ref{fig: pre-training} in Section~\ref{imp: pre}. Then, we demonstrate details few-shot experiments on classification and semantic segmentation tasks on Agriculture-Vision shown in the right part of Figure \ref{fig: fine-tune} in Section~\ref{imp: AG_cls}. Following this, we list the key parameters for valuation in the few-shot semantic segmentation task in Section~\ref{imp: AG_seg}. Lastly, we report the necessary details for EuroSAT in Section \label{imp: SAT_cls}.

\subsubsection{Pre-training on Extended Agriculture-Vision}
\label{imp: pre}
GenCo uses different ResNet as its encoder and two layers of MLP as a projector as the basis for our contrastive learning framework. The following generator $G$ is instantiated with three linear layers and a ReLU layer between two successive layers. The output feature vectors, i.e., $q$, $q'$, and $k$ have dimensions of 128, and there are 16,384 negative keys stored in the memory bank. We pre-train the generator-based contrastive learning model for 200 epochs using an SGD optimizer, a learning rate of 0.3, and a weight decay of 0.0001. The learning rate is adjusted to 0.03 and 0.003 at the 140th and 160th epochs accordingly. As this data is hyperspectral, we add one more channel to the encoder during the training for the NIR input, and this extra channel is initialized with the same weights as the red channel.

 Besides the proposed GenCo, we also conduct experiments using MoCo-V2\cite{chen2020improved}, SimSiam\cite{chen2021exploring} and SimCLR\cite{chen2020simple} as alternative methods for pre-training. Results are also reported in the following sections. For MoCo-V2, we train the model with the exact same hyperparameters as the proposed method, given the similarity of the structure. For SimSiam, we train the models for 200 epochs with a batch size of 256. The learning rate = 0.05 with an SGD optimizer. The weight decay is 0.0001 and the SGD momentum is 0.9. For pre-training of SimCLR, we reproduce SimCLR with a smaller batch size of 512 and cosine appealed learning rate of 0.05. We follow the self-supervised learning paradigm in \cite{chen2020big} without distillation for a fair comparison of the other two contrastive learning methods. All the experiments are conducted on a server with 8 GPUs.

\subsubsection{Few-Shot Classification on Agriculture-Vision}
\label{imp: AG_cls}
The first set of experiments focuses on the classification formulation of the Agriculture-Vision task.  We use ResNet-18, ResNet-50, and ResNet-101 as the backbones for fine-tuning. All backbones are fixed except the last fully connected layer and the generator $G$, which are learnable. Different from the optimization methods used in \cite{he2020momentum}, we use Adam as an optimizer for all experiments with an initial learning rate set to 0.001. We train the classification models for 100 epochs with a batch size of 64.

\subsubsection{Few-Shot Segmentation on Agriculture-Vision}
\label{imp: AG_seg}
Following the work from \cite{chiu2020agriculture}, we ignore storm damage annotations when performing evaluations due to their extreme scarcity. Similar to the fine-tuning strategy we used in the classification task, we freeze all backbones during training but with a learnable five-layer decoder. The decoders are randomly initialized and attached to the encoders, forming a lightweight and imbalanced U-Net. We use the AdamW optimizer with the learning rate set to 6e-5 and the one learning rate cycle scheduler proposed by \cite{smith2019super}. In total, we train the segmentation models for 100 epochs with 300 steps per epoch. For all experiments, we use a batch size of 64 during fine-tuning. 

\subsubsection{Few-Shot Classification on EuroSAT}
\label{imp: SAT_cls}
We additionally illustrate that embeddings learned from GenCo on AV+ help few-shot learning tasks in the more general remote sensing community. To achieve this, we evaluate our proposed method on the few-shot classification task of EuroSAT \cite{helber2019eurosat}. Following experiments in previous sections, we evaluate the quality and adaptability of pre-trained features from the proposed methods in this land-cover classification task.

We add one fully connected layer to pre-trained backbones, building the classifier for EuroSAT. We train the model for 100 epochs with an AdamW optimizer and a batch size of 256. The initial learning rate is 0.001.

\begin{table}

    \centering
    \caption{Comparison of fine-tuning results between weights from supervised ImageNet and weights from GenCo on AV+ for the 10-shot classification}
    \begin{tabular}{c | l | l }
    \toprule
        Pre-trained Weight & Backbone & Accuracy(\%) \\ 
    \midrule
        Sup. on ImageNet  & ResNet-18 & 55.22 $ \pm $0.34 \\  
        SimSiam on AV+ & ResNet-18 & 63.12 $ \pm $0.71\\  
        SimCLR on AV+ & ResNet-18 & 63.59 $ \pm $0.83\\  
        MoCoV2 on AV+ & ResNet-18 & 64.17 $ \pm $0.62\\        
        GenCo on AV+ & ResNet-18 & \textbf{65.51} $ \pm $0.68 \\ \hline  
        ImageNet & ResNet-50 & 56.53 $ \pm $0.43 \\  
        SimSiam on AV+ & ResNet-50 & 62.41 $ \pm $0.78\\  
        SimCLR on AV+ & ResNet-50 & 63.55 $ \pm $0.77 \\  
        MoCoV2 on AV+ & ResNet-50 & 63.80 $ \pm $0.72\\  
        GenCo on AV+  & ResNet-50 & \textbf{64.82} $ \pm $0.70\\ \hline  
        Sup. on ImageNet & ResNet-101 & 54.34 $ \pm $0.45\\  
        SimSiam on AV+ & ResNet-101 & 62.27 $ \pm $0.89\\  
        SimCLR on AV+ & ResNet-101 & 63.50 $ \pm $0.80\\  
        MoCoV2 on AV+ & ResNet-101 & 64.12 $ \pm $0.78\\  
        GenCo on AV+ & ResNet-101 & \textbf{64.62} $ \pm $0.49\\ 
    \bottomrule
    \end{tabular}
    \label{TAB: classify AG1}
      \vspace*{-5mm} 
\end{table}

\subsection{Results of Experiments}
\label{sec: results}
\subsubsection{Few-shot Learning on Agriculture-Vision}

\textbf{Quality of Pre-Trained Embeddings.} We first prove the quality and adaptability of pre-trained embeddings from the proposed methods. As shown in Table~\ref{TAB: classify AG1}, our pre-trained weights show significantly better results than those from ImageNet, with over 10 points improvement on average. These results prove the adaptability of embeddings encoded from our pre-trained weights and better generalization capacity in this few-shot classification task for agricultural patterns. The best result is obtained from ResNet-18 instead of the larger ResNet-50 or ResNet-101. With only 10 shots, this observation is due to the last layer attached to ResNet-18 being smaller than the fully connected layers in larger backbones. MoCoV2, SimSiam and SimCLR show sub-optimal results compared with GenCo. However, with our two-stage training strategy, all AV+ pre-trained weights enable better performance than any ImageNet weights.

\begin{table}
    \centering
    \caption{Comparison of the classification task between the 10-shot results of GenCo and end-to-end training using the full Agriculture-Vision on ResNet-18.}
    \begin{tabular}{c | l | l | l }
    \toprule
        \makecell[c]{Pre-trained \\ Weight} & \makecell[c]{Freeze \\ Backbone}   & \makecell[c]{Number of \\Images} & \makecell[c]{Accuracy\\ (\%)} \\ 
    \midrule
        Random & False & 9,000 &57.30 $ \pm $0.81 \\ \hline
        Random & False & 94,986 & 62.31 $ \pm $0.25 \\ \hline
        GenCo on AV+ & True & \textbf{10} &   \textbf{65.51}$ \pm $0.68\\ 
    \bottomrule
    \end{tabular}
    \label{TAB: classify AG2}
  \vspace*{-5mm} 
\end{table}

\begin{table}
    \centering
    \caption{9-way few-shot classification accuracy on Agriculture-Vision based on weights pre-trained from GenCo.}
    \begin{tabular}{c|l|l|l }
    \toprule
         Backbone $\setminus$  Shots & \makecell[c]{Accuracy\\10 shots\\(\%)}& \makecell[c]{Accuracy\\5 shots\\(\%)} & \makecell[c]{Accuracy\\1 shot\\(\%)}\\ 
    \midrule
        ResNet-18 & \textbf{65.51} $ \pm $0.68  & \textbf{61.61} $ \pm $0.71  & 16.72  $ \pm $1.32\\ \hline
        ResNet-50 & 64.82 $ \pm $0.70 & 59.44 $ \pm $0.79 & \textbf{29.64} $ \pm $1.28 \\ \hline
        ResNet-101 & 64.62 $ \pm $0.49 & 59.56$ \pm $0.66  & 28.84 $ \pm $1.73 \\ 
    \bottomrule
    \end{tabular}
    \label{Tab: classify AG3}
      \vspace*{-5mm} 
\end{table}

\begin{table}
    \centering
    \caption{Comparison of fine-tuning results between weights pre-trained on COCO and weights from GenCo on AV+ for the 10-shot semantic segmentation task.}
    \begin{tabular}{c|l|l}
    \toprule

       Pre-trained Weight & Backbone & \makecell[c]{mIoU - 8 Classes} \\ 
    \midrule
        COCO & ResNet-18 & 15.61 \\ 
        GenCo on AV+ & ResNet-18 & \textbf{23.56} \\ \hline 
        COCO & ResNet-50 & 15.60 \\ 
        GenCo on AV+ & ResNet-50 & \textbf{23.00} \\ \hline 
        COCO & ResNet-101 & 15.19 \\ 
        GenCo on AV+ & ResNet-101 & \textbf{21.04} \\ 
    \bottomrule
    \end{tabular}
    \label{TAB: Seg1}
      \vspace*{-2mm} 
\end{table}

\begin{table*}
    \centering
    \caption{Comparison of the segmentation task between the 10-shot results of the proposed method and end-to-end training using the full Agriculture-Vision on ResNet-18 and ResNet-50.}
    \begin{tabular}{c |l|l|l|l}
    \toprule
        Pre-trained Weight & Backbone & Freeze Backbone & Number of Images & mIoU - 8 Classes \\ 
    \midrule
        Random & ResNet-18 & False & 9000 & 19.02 \\ 
        Random & ResNet-18 & False & 94986 & 21.37 \\ 
        SimSiam on AV+     & ResNet-18 & True & \textbf{10} & 21.30 \\
        SimCLR on AV+     & ResNet-18 & True & \textbf{10} & 22.13 \\
        MoCo on AV+     & ResNet-18 & True & \textbf{10} & 22.11 \\
        GenCo on AV+  & ResNet-18 & True & \textbf{10} & \textbf{23.56} \\ \hline 
        Random & ResNet-50 & False & 9000 & 19.58 \\ 
        Random & ResNet-50 & False & 94986 & 21.82 \\ 
        SimSiam on AV+     & ResNet-50 & True & \textbf{10} & 21.06 \\
        SimCLR on AV+     & ResNet-50 & True & \textbf{10} & 21.98 \\
        MoCo on AV+     & ResNet-50 & True & \textbf{10} & 21.21 \\
        GenCo on AV+  & ResNet-50 & True & \textbf{10} & \textbf{23.00} \\
    \bottomrule
    \end{tabular}
    \label{TAB: Seg2}

\end{table*}

\textbf{Learning Efficiency.} Next, we continue to demonstrate the learning efficiency of GenCo by comparing it with the model trained with 94,986 labeled images. For models training in an end-to-end manner, there is a noticeable drop once we reduce the number of models for training. However, as shown in Table~\ref{TAB: classify AG2}, GenCo outperforms model training with numerous images with little computation and much fewer labels for agricultural pattern classification. This observation is important as it illustrates the potential of training diverse  deep-learning tasks in agriculture and remote sensing with minimum effort but still providing satisfactory results.

Table \ref{Tab: classify AG3} demonstrates the 9-way few-shot classification results with different sizes of backbones. All results are averaged from 3 trials and use the same training setup for a fair comparison. While ResNet-18 gives the best results when trained with five shots or ten shots, ResNet-50 shows the best results when there is only one labeled sample for each class. The performance of ResNet-50 and ResNet-101 are very similar. Generally, favorable results can be acquired when the number of shots is five or greater.

\subsubsection{Ablation Study of Few-shot Classification}

In this section, we aim to prove the effectiveness of the proposed generator $G$ in the contrastive learning framework. We conduct experiments on GenCo with ResNet-18 as the backbones for pre-training on AV+. We then report the results of the downstream 10-shot classification tasks on the agriculture vision dataset.

\begin{table}
  \vspace*{-5mm} 
    \centering
    \caption{
    Ablation of the proposed generator with different contrastive learning frameworks.
    }
    \begin{tabular}{ l | ccc}
    \hline
    \toprule
    {Shots $\setminus$  Modules}  &{\makecell[c]{No Generator\\ (MoCo-V2)}} &{\makecell[c]{Generator with \\Pre-training }}   &{\makecell[c]{GenCo}}  \\

    \midrule

    10 shot       & 64.17 &64.61 & \textbf{65.51} \\
     5  shot       & 59.32 &59.91 &\textbf{61.61} \\
    1 shot       &15.21 &15.23 & \textbf{16.72} \\
    \bottomrule
    \end{tabular}
    \label{ablation: modules}
 
\end{table}

We conducted a comparative analysis of three frameworks: MoCo-V2, MoCo-V2 with a generator, and our proposed GenCo. Our experimental results, as summarized in Table~\ref{ablation: modules}, demonstrate that the use of a generator improves the learned embeddings even when introduced solely during pre-training. This improvement can be attributed to the additional variance and contrast introduced by the positive feature vectors. However, the performance gain achieved is relatively marginal. Notably, the largest gain occurs when we adopt the pre-trained generator to few-shot learning, which provides additional labeled data in the feature space. As such, the generator can enhance both pre-training and downstream tasks simultaneously, thereby improving the overall performance of the framework.

\subsubsection{Few-shot Segmentation on Agriculture-Vision}
\textbf{Quality of Pre-Trained Embeddings.} Since U-Net's structures contain skip connections from different layers  \cite{ronneberger2015u}, we don't evaluate a single embedding but features from different scales. Concretely, features from GenCo and features from encoders pre-trained on COCO are compared using the mean intersection over union (mIoU) metric.  As reported in Table~\ref{TAB: Seg1}, our proposed method shows around 6-8 points of improvement compared with weights pre-trained on COCO. Consistent with the results from the classification task, the best mIoU is reached by ResNet-18 with a smaller decoder attached. The other conclusion we can draw is that the feature distribution pre-trained from natural images (COCO) and remote sensing images (AV+) is significantly different. Therefore, we can observe a noticeable improvement in the results pre-trained on AV+. 

\textbf{Learning Efficiency.} We also examine the learning efficiency of the segmentation task. To do this, we use only ten sampled images per category and compare the results of GenCo with those of models trained on the full Agriculture-Vision dataset. While our approach fixes the backbone, we unfreeze the segmentation model's encoder and the generator training on the full dataset. Based on the results presented in Table \ref{TAB: Seg2}, we observe an improvement of 2.19 points and 1.18 points for ResNet-18 and ResNet-50, respectively, using the GenCo approach. However, for MoVo-V2, SimSiam, and SimCLR, the results are comparable or only slightly better than those obtained using the end-to-end training method. Importantly, while the GenCo-based few-shot segmentation approach still outperforms models trained with a large number of labeled images, we note that the improvement is not as significant as the improvement achieved in the classification task. This observation is likely because the decoders used for segmentation have more parameters to be tuned than a single-layer classifier. With limited labeled samples, smaller models are better able to avoid overfitting and show better results. Therefore, in this few-shot segmentation task, ResNet-18 performs the most satisfactorily.

\subsubsection{Few-shot Classification on EuroSAT}
\textbf{Quality of Pre-Trained Embeddings.} Results show that GenCo still leads to better embeddings on this remote sensing dataset. As seen in Table~\ref{TAB: Euro1}, features from GenCo improve one percent of accuracy on average compared to the features trained from ImageNet. Since EuroSAT shares much less similarity with our pre-trained dataset, i.e., AV+, the improvement is moderate. However, the gain is still stably earned, crossing different sizes of backbones. The results from MoCo-V2, SimSiam and SimCLR still outperform the results of ImageNet but are sub-optimal compared with results from GenCo, proving the effectiveness of the proposed generator.

\begin{table}
  \vspace*{-5mm} 
    \centering
    \caption{Comparison of fine-tuning results between weights pre-trained on supervised ImageNet and weights from our GenCo on EuroSAT for the 10-shot classification}
    \begin{tabular}{c |l | l }
    \toprule
        Pre-trained Weight & Backbone & Accuracy(\%) \\  
    \midrule
        ImageNet  & ResNet-18 & 66.90 $ \pm $0.11\\  
        SimSiam on AV+ & ResNet-18 & 67.20 $ \pm $0.43\\  
        SimCLR on AV+ & ResNet-18 & 67.31 $ \pm $0.51\\  
        MoCo on AV+ & ResNet-18 & 67.14 $ \pm $0.32\\          
        GenCo on AV+ & ResNet-18 & \textbf{67.92}$ \pm $0.31 \\  \hline  
        ImageNet  & ResNet-50 & 65.01 $ \pm $0.14\\ 
        SimSiam on AV+ & ResNet-50 &  {65.61} $ \pm $0.38\\   
        SimCLR on AV+ & ResNet-50 & 65.93 $ \pm $0.62\\  
        MoCo on AV+ & ResNet-50 & 65.58 $ \pm $0.40\\  
        GenCo on AV+  & ResNet-50 & \textbf{66.11} $ \pm $0.45\\ \hline  
        ImageNet   & ResNet-101 & 63.34 $ \pm $0.20\\  
        SimSiam on AV+ & ResNet-101 &  {63.32} $ \pm $0.35\\ 
        SimCLR on AV+ & ResNet-101 & 64.17 $ \pm $0.68\\  
        MoCo on AV+ & ResNet-101 & 63.79 $ \pm $0.45\\  
        GenCo on AV+  & ResNet-101 & \textbf{64.79} $ \pm $0.46\\ 
    \bottomrule
    \end{tabular}
    \label{TAB: Euro1}
  \vspace*{-5mm} 
\end{table}

\begin{table}
    \centering
    \caption{Comparison of the classification task between the 10-shot results of GenCo and end-to-end training using the full EuroSAT on ResNet-18. *: results referred from \cite{manas2021seasonal}} 
    \begin{tabular}{c|l|l|l}
    \toprule
        \makecell[c]{Pre-trained \\ Weight} & \makecell[c]{Freeze \\ Backbone}   & \makecell[c]{Number of \\Images} & \makecell[c]{Accuracy\\ (\%)} \\ 
     \midrule       
        Random & False & 2,700 & 58.81$ \pm $0.10 \\ 
        Random & False & 27,000 & 63.21 *\\ 
        Random & False & 27,000 & 63.34 $ \pm $0.08\\ \hline
        GenCo on AV+ & True & 10 & \textbf{66.90} $ \pm $0.31 \\ 
    \bottomrule
    \end{tabular}
    \label{TAB: Euro2}
  \vspace*{-2mm} 
\end{table}

\textbf{Learning Efficiency.} In the experiments on label efficiency, we continue to compare the few-shot classification models with those models randomly initialized and trained on 27,000 labeled images. The proposed method outperforms the end-to-end model by 3.66 points in this classification task with only ten labeled images, as shown in Table~\ref{TAB: Euro2}. This result is crucial as it proves the effectiveness of our methods in different domains. With a remarkably cheap effort of labeling, it re-verifies the vast possibility of deploying our models to various downstream tasks in agriculture and remote sensing.

\begin{table}
  \vspace*{-3mm} 
    \centering
    \caption{10-way few-shot classification accuracy on EuroSAT based on weights pre-trained from GenCo.}
    \begin{tabular}{c|l|l|l }
    \toprule
         Backbone $\setminus$  Shots & \makecell[c]{Accuracy\\10 shots\\(\%)}& \makecell[c]{Accuracy\\5 shots\\(\%)} & \makecell[c]{Accuracy\\1 shot\\(\%)}\\ 
    \midrule
        ResNet-18 & \textbf{67.92}$ \pm $0.31  & \textbf{63.20}$ \pm $0.37  & \textbf{11.50}$ \pm $0.82  \\ \hline
        ResNet-50 & 65.01$ \pm $0.45  & 59.40$ \pm $0.51  & 11.21$ \pm $1.42  \\ \hline
        ResNet-101 & 63.70$ \pm $0.46  & 58.70 $ \pm $0.66 & 11.40$ \pm $1.71  \\ 
    \bottomrule
    \end{tabular}
    \label{Tab: Euro3}
\end{table}

We show the results of the 10-way few-shot classification on EuroSAT in the following Table \ref{Tab: Euro3}. For a complete and fair comparison, we report the performance of backbones with different sizes and average results over experiments with three random seeds. More specifically, the ResNet-18 shows the most satisfactory performance crossing various backbones and shots. While we can notice a 1 to 4 points drop in accuracy when we increase the size of backbones under 10 or 5 shots settings, one-shot classification shows very similar performance regardless of encoder sizes. 
\section{Conclusion}
In this study, we proposed GenCo, a generator-based two-stage approach for few-shot classification and segmentation on remote sensing and earth observation data. Our method proved to be effective due to the sufficient contrast introduced by the generator during pre-training and the remarkable adaptability of embeddings in downstream few-shot tasks. Furthermore, we demonstrated that the generator could be integrated into the few-shot learning framework to further address the issue of data scarcity and enrich the learning information. Most importantly, our approach provides an alternative solution to the labeling challenges in agriculture and remote sensing domains. With our few-shot contrastive learning-based approach, we believe that it is possible to deploy models in the real world with minimal labeled data and training effort. 



\bibliography{ecai}

\begin{thebibliography}{10}

\bibitem{alajaji2020few}
Dalal Alajaji, Haikel~S Alhichri, Nassim Ammour, and Naif Alajlan, `Few-shot
  learning for remote sensing scene classification', in {\em 2020 Mediterranean
  and Middle-East Geoscience and Remote Sensing Symposium (M2GARSS)}, pp.
  81--84. IEEE, (2020).

\bibitem{alajaji2020meta}
Dalal~A Alajaji and Haikel Alhichri, `Few shot scene classification in remote
  sensing using meta-agnostic machine', in {\em 2020 6th conference on data
  science and machine learning applications (CDMA)}, pp. 77--80. IEEE, (2020).

\bibitem{cacheux2019modeling}
Yannick~Le Cacheux, Herve~Le Borgne, and Michel Crucianu, `Modeling inter and
  intra-class relations in the triplet loss for zero-shot learning', in {\em
  Proceedings of the IEEE/CVF International Conference on Computer Vision}, pp.
  10333--10342, (2019).

\bibitem{chen2019claims}
Suiyao Chen, Nan Kong, Xuxue Sun, Hongdao Meng, and Mingyang Li, `Claims
  data-driven modeling of hospital time-to-readmission risk with latent
  heterogeneity', {\em Health care management science}, {\bf 22},  156--179,
  (2019).

\bibitem{chen2020simple}
Ting Chen, Simon Kornblith, Mohammad Norouzi, and Geoffrey Hinton, `A simple
  framework for contrastive learning of visual representations', in {\em
  International conference on machine learning}, pp. 1597--1607. PMLR, (2020).

\bibitem{chen2020big}
Ting Chen, Simon Kornblith, Kevin Swersky, Mohammad Norouzi, and Geoffrey~E
  Hinton, `Big self-supervised models are strong semi-supervised learners',
  {\em Advances in neural information processing systems}, {\bf 33},
  22243--22255, (2020).

\bibitem{chen2021momentum}
Xiaocong Chen, Lina Yao, Tao Zhou, Jinming Dong, and Yu~Zhang, `Momentum
  contrastive learning for few-shot covid-19 diagnosis from chest ct images',
  {\em Pattern recognition}, {\bf 113},  107826, (2021).

\bibitem{chen2020improved}
Xinlei Chen, Haoqi Fan, Ross Girshick, and Kaiming He, `Improved baselines with
  momentum contrastive learning', {\em arXiv preprint arXiv:2003.04297},
  (2020).

\bibitem{chen2021exploring}
Xinlei Chen and Kaiming He, `Exploring simple siamese representation learning',
  in {\em Proceedings of the IEEE/CVF Conference on Computer Vision and Pattern
  Recognition}, pp. 15750--15758, (2021).

\bibitem{chen2019image}
Zitian Chen, Yanwei Fu, Yu-Xiong Wang, Lin Ma, Wei Liu, and Martial Hebert,
  `Image deformation meta-networks for one-shot learning', in {\em Proceedings
  of the IEEE/CVF conference on computer vision and pattern recognition}, pp.
  8680--8689, (2019).

\bibitem{Chiu_2020_CVPR_Workshops}
Mang~Tik Chiu, Xingqian Xu, Kai Wang, Jennifer Hobbs, Naira Hovakimyan,
  Thomas~S. Huang, and Honghui Shi, `The 1st agriculture-vision challenge:
  Methods and results', in {\em Proceedings of the IEEE/CVF Conference on
  Computer Vision and Pattern Recognition (CVPR) Workshops}, (June 2020).

\bibitem{chiu2020agriculture}
Mang~Tik Chiu, Xingqian Xu, Yunchao Wei, Zilong Huang, Alexander~G Schwing,
  Robert Brunner, Hrant Khachatrian, Hovnatan Karapetyan, Ivan Dozier, Greg
  Rose, et~al., `Agriculture-vision: A large aerial image database for
  agricultural pattern analysis', in {\em Proceedings of the IEEE/CVF
  Conference on Computer Vision and Pattern Recognition}, pp. 2828--2838,
  (2020).

\bibitem{deng2009imagenet}
Jia Deng, Wei Dong, Richard Socher, Li-Jia Li, Kai Li, and Li~Fei-Fei,
  `Imagenet: A large-scale hierarchical image database', in {\em 2009 IEEE
  conference on computer vision and pattern recognition}, pp. 248--255. Ieee,
  (2009).

\bibitem{doersch2020crosstransformers}
Carl Doersch, Ankush Gupta, and Andrew Zisserman, `Crosstransformers:
  spatially-aware few-shot transfer', {\em Advances in Neural Information
  Processing Systems}, {\bf 33},  21981--21993, (2020).

\bibitem{fei2006one}
Li~Fei-Fei, Robert Fergus, and Pietro Perona, `One-shot learning of object
  categories', {\em IEEE transactions on pattern analysis and machine
  intelligence}, {\bf 28}(4),  594--611, (2006).

\bibitem{finn2017model}
Chelsea Finn, Pieter Abbeel, and Sergey Levine, `Model-agnostic meta-learning
  for fast adaptation of deep networks', in {\em International conference on
  machine learning}, pp. 1126--1135. PMLR, (2017).

\bibitem{finn2017one}
Chelsea Finn, Tianhe Yu, Tianhao Zhang, Pieter Abbeel, and Sergey Levine,
  `One-shot visual imitation learning via meta-learning', in {\em Conference on
  robot learning}, pp. 357--368. PMLR, (2017).

\bibitem{gidaris2019boosting}
Spyros Gidaris, Andrei Bursuc, Nikos Komodakis, Patrick P{\'e}rez, and Matthieu
  Cord, `Boosting few-shot visual learning with self-supervision', in {\em
  Proceedings of the IEEE/CVF international conference on computer vision}, pp.
  8059--8068, (2019).

\bibitem{gidaris2018dynamic}
Spyros Gidaris and Nikos Komodakis, `Dynamic few-shot visual learning without
  forgetting', in {\em Proceedings of the IEEE conference on computer vision
  and pattern recognition}, pp. 4367--4375, (2018).

\bibitem{gidaris2019generating}
Spyros Gidaris and Nikos Komodakis, `Generating classification weights with gnn
  denoising autoencoders for few-shot learning', in {\em Proceedings of the
  IEEE/CVF conference on computer vision and pattern recognition}, pp. 21--30,
  (2019).

\bibitem{grill2020bootstrap}
Jean-Bastien Grill, Florian Strub, Florent Altch{\'e}, Corentin Tallec, Pierre
  Richemond, Elena Buchatskaya, Carl Doersch, Bernardo Avila~Pires, Zhaohan
  Guo, Mohammad Gheshlaghi~Azar, et~al., `Bootstrap your own latent-a new
  approach to self-supervised learning', {\em Advances in neural information
  processing systems}, {\bf 33},  21271--21284, (2020).

\bibitem{he2020momentum}
Kaiming He, Haoqi Fan, Yuxin Wu, Saining Xie, and Ross Girshick, `Momentum
  contrast for unsupervised visual representation learning', in {\em
  Proceedings of the IEEE/CVF conference on computer vision and pattern
  recognition}, pp. 9729--9738, (2020).

\bibitem{helber2019eurosat}
Patrick Helber, Benjamin Bischke, Andreas Dengel, and Damian Borth, `Eurosat: A
  novel dataset and deep learning benchmark for land use and land cover
  classification', {\em IEEE Journal of Selected Topics in Applied Earth
  Observations and Remote Sensing}, {\bf 12}(7),  2217--2226, (2019).

\bibitem{kemker2018low}
Ronald Kemker, Ryan Luu, and Christopher Kanan, `Low-shot learning for the
  semantic segmentation of remote sensing imagery', {\em IEEE Transactions on
  Geoscience and Remote Sensing}, {\bf 56}(10),  6214--6223, (2018).

\bibitem{koch2015siamese}
Gregory Koch, Richard Zemel, Ruslan Salakhutdinov, et~al., `Siamese neural
  networks for one-shot image recognition', in {\em ICML deep learning
  workshop}, volume~2, p.~0. Lille, (2015).

\bibitem{lee2019meta}
Kwonjoon Lee, Subhransu Maji, Avinash Ravichandran, and Stefano Soatto,
  `Meta-learning with differentiable convex optimization', in {\em Proceedings
  of the IEEE/CVF conference on computer vision and pattern recognition}, pp.
  10657--10665, (2019).

\bibitem{li2022global}
Haifeng Li, Yi~Li, Guo Zhang, Ruoyun Liu, Haozhe Huang, Qing Zhu, and Chao Tao,
  `Global and local contrastive self-supervised learning for semantic
  segmentation of hr remote sensing images', {\em IEEE Transactions on
  Geoscience and Remote Sensing}, {\bf 60},  1--14, (2022).

\bibitem{li2020adversarial}
Kai Li, Yulun Zhang, Kunpeng Li, and Yun Fu, `Adversarial feature hallucination
  networks for few-shot learning', in {\em Proceedings of the IEEE/CVF
  Conference on Computer Vision and Pattern Recognition}, pp. 13470--13479,
  (2020).

\bibitem{li2020dla}
Lingjun Li, Junwei Han, Xiwen Yao, Gong Cheng, and Lei Guo, `Dla-matchnet for
  few-shot remote sensing image scene classification', {\em IEEE Transactions
  on Geoscience and Remote Sensing}, {\bf 59}(9),  7844--7853, (2020).

\bibitem{li2021few}
Xiang Li, Jingyu Deng, and Yi~Fang, `Few-shot object detection on remote
  sensing images', {\em IEEE Transactions on Geoscience and Remote Sensing},
  {\bf 60},  1--14, (2021).

\bibitem{lin2014microsoft}
Tsung-Yi Lin, Michael Maire, Serge Belongie, James Hays, Pietro Perona, Deva
  Ramanan, Piotr Doll{\'a}r, and C~Lawrence Zitnick, `Microsoft coco: Common
  objects in context', in {\em European conference on computer vision}, pp.
  740--755. Springer, (2014).

\bibitem{liu2018deep}
Bing Liu, Xuchu Yu, Anzhu Yu, Pengqiang Zhang, Gang Wan, and Ruirui Wang, `Deep
  few-shot learning for hyperspectral image classification', {\em IEEE
  Transactions on Geoscience and Remote Sensing}, {\bf 57}(4),  2290--2304,
  (2018).

\bibitem{luo2017thinet}
Jian-Hao Luo, Jianxin Wu, and Weiyao Lin, `Thinet: A filter level pruning
  method for deep neural network compression', in {\em Proceedings of the IEEE
  international conference on computer vision}, pp. 5058--5066, (2017).

\bibitem{manas2021seasonal}
Oscar Manas, Alexandre Lacoste, Xavier Gir{\'o}-i Nieto, David Vazquez, and Pau
  Rodriguez, `Seasonal contrast: Unsupervised pre-training from uncurated
  remote sensing data', in {\em Proceedings of the IEEE/CVF International
  Conference on Computer Vision}, pp. 9414--9423, (2021).

\bibitem{nichol2018reptile}
Alex Nichol and John Schulman, `Reptile: a scalable metalearning algorithm',
  {\em arXiv preprint arXiv:1803.02999}, {\bf 2}(3), ~4, (2018).

\bibitem{oord2018representation}
Aaron van~den Oord, Yazhe Li, and Oriol Vinyals, `Representation learning with
  contrastive predictive coding', {\em arXiv preprint arXiv:1807.03748},
  (2018).

\bibitem{street2sat}
Madhava Paliyam, Catherine~L Nakalembe, Kevin Liu, Richard Nyiawung, and
  Hannah~R Kerner, `Street2sat: A machine learning pipeline for generating
  ground-truth geo-referenced labeled datasets from street-level images', in
  {\em ICML 2021 Workshop on Tackling Climate Change with Machine Learning},
  (2021).

\bibitem{qi2018low}
Hang Qi, Matthew Brown, and David~G Lowe, `Low-shot learning with imprinted
  weights', in {\em Proceedings of the IEEE conference on computer vision and
  pattern recognition}, pp. 5822--5830, (2018).

\bibitem{ronneberger2015u}
Olaf Ronneberger, Philipp Fischer, and Thomas Brox, `U-net: Convolutional
  networks for biomedical image segmentation', in {\em International Conference
  on Medical image computing and computer-assisted intervention}, pp. 234--241.
  Springer, (2015).

\bibitem{smith2019super}
Leslie~N Smith and Nicholay Topin, `Super-convergence: Very fast training of
  neural networks using large learning rates', in {\em Artificial intelligence
  and machine learning for multi-domain operations applications}, volume 11006,
  pp. 369--386. SPIE, (2019).

\bibitem{su2020does}
Jong-Chyi Su, Subhransu Maji, and Bharath Hariharan, `When does
  self-supervision improve few-shot learning?', in {\em European conference on
  computer vision}, pp. 645--666. Springer, (2020).

\bibitem{sun2021research}
Xian Sun, Bing Wang, Zhirui Wang, Hao Li, Hengchao Li, and Kun Fu, `Research
  progress on few-shot learning for remote sensing image interpretation', {\em
  IEEE Journal of Selected Topics in Applied Earth Observations and Remote
  Sensing}, {\bf 14},  2387--2402, (2021).

\bibitem{sung2018learning}
Flood Sung, Yongxin Yang, Li~Zhang, Tao Xiang, Philip~HS Torr, and Timothy~M
  Hospedales, `Learning to compare: Relation network for few-shot learning', in
  {\em Proceedings of the IEEE conference on computer vision and pattern
  recognition}, pp. 1199--1208, (2018).

\bibitem{tao2022optimizing}
Ran Tao, Pan Zhao, Jing Wu, Nicolas~F Martin, Matthew~T Harrison, Carla
  Ferreira, Zahra Kalantari, and Naira Hovakimyan, `Optimizing crop management
  with reinforcement learning and imitation learning', {\em arXiv preprint
  arXiv:2209.09991}, (2022).

\bibitem{verma2019manifold}
Vikas Verma, Alex Lamb, Christopher Beckham, Amir Najafi, Ioannis Mitliagkas,
  David Lopez-Paz, and Yoshua Bengio, `Manifold mixup: Better representations
  by interpolating hidden states', in {\em International Conference on Machine
  Learning}, pp. 6438--6447. PMLR, (2019).

\bibitem{vinyals2016matching}
Oriol Vinyals, Charles Blundell, Timothy Lillicrap, Daan Wierstra, et~al.,
  `Matching networks for one shot learning', {\em Advances in neural
  information processing systems}, {\bf 29}, (2016).

\bibitem{wang2021dmml}
Bing Wang, Zhirui Wang, Xian Sun, Hongqi Wang, and Kun Fu, `Dmml-net: Deep
  metametric learning for few-shot geographic object segmentation in remote
  sensing imagery', {\em IEEE Transactions on Geoscience and Remote Sensing},
  {\bf 60},  1--18, (2021).

\bibitem{wang2022global}
Haoxiang Wang, Yite Wang, Ruoyu Sun, and Bo~Li, `Global convergence of maml and
  theory-inspired neural architecture search for few-shot learning', in {\em
  Proceedings of the IEEE/CVF Conference on Computer Vision and Pattern
  Recognition}, pp. 9797--9808, (2022).

\bibitem{wang2019panet}
Kaixin Wang, Jun~Hao Liew, Yingtian Zou, Daquan Zhou, and Jiashi Feng, `Panet:
  Few-shot image semantic segmentation with prototype alignment', in {\em
  Proceedings of the IEEE/CVF International Conference on Computer Vision}, pp.
  9197--9206, (2019).

\bibitem{wang2020frustratingly}
Xin Wang, Thomas~E Huang, Trevor Darrell, Joseph~E Gonzalez, and Fisher Yu,
  `Frustratingly simple few-shot object detection', {\em arXiv preprint
  arXiv:2003.06957}, (2020).

\bibitem{wang2023double}
Yite Wang, Jing Wu, Naira Hovakimyan, and Ruoyu Sun, `Double dynamic sparse
  training for gans', {\em arXiv preprint arXiv:2302.14670}, (2023).

\bibitem{wang2018low}
Yu-Xiong Wang, Ross Girshick, Martial Hebert, and Bharath Hariharan, `Low-shot
  learning from imaginary data', in {\em Proceedings of the IEEE conference on
  computer vision and pattern recognition}, pp. 7278--7286, (2018).

\bibitem{wu2023hallucination}
Jing Wu, Jennifer Hobbs, and Naira Hovakimyan, `Hallucination improves the
  performance of unsupervised visual representation learning', {\em arXiv
  preprint arXiv:2307.12168}, (2023).

\bibitem{wu2023extended}
Jing Wu, David Pichler, Daniel Marley, David Wilson, Naira Hovakimyan, and
  Jennifer Hobbs, `Extended agriculture-vision: An extension of a large aerial
  image dataset for agricultural pattern analysis', {\em arXiv preprint
  arXiv:2303.02460}, (2023).

\bibitem{wu2022optimizing}
Jing Wu, Ran Tao, Pan Zhao, Nicolas~F Martin, and Naira Hovakimyan, `Optimizing
  nitrogen management with deep reinforcement learning and crop simulations',
  in {\em Proceedings of the IEEE/CVF Conference on Computer Vision and Pattern
  Recognition}, pp. 1712--1720, (2022).

\bibitem{yang2022survey}
Jiachen Yang, Xiaolan Guo, Yang Li, Francesco Marinello, Sezai Ercisli, and
  Zhuo Zhang, `A survey of few-shot learning in smart agriculture:
  developments, applications, and challenges', {\em Plant Methods}, {\bf
  18}(1),  1--12, (2022).

\bibitem{yao2021scale}
Xiwen Yao, Qinglong Cao, Xiaoxu Feng, Gong Cheng, and Junwei Han, `Scale-aware
  detailed matching for few-shot aerial image semantic segmentation', {\em IEEE
  Transactions on Geoscience and Remote Sensing}, {\bf 60},  1--11, (2021).

\bibitem{zbontar2021barlow}
Jure Zbontar, Li~Jing, Ishan Misra, Yann LeCun, and St{\'e}phane Deny, `Barlow
  twins: Self-supervised learning via redundancy reduction', in {\em
  International Conference on Machine Learning}, pp. 12310--12320. PMLR,
  (2021).

\bibitem{zeng2022task}
Qingjie Zeng and Jie Geng, `Task-specific contrastive learning for few-shot
  remote sensing image scene classification', {\em ISPRS Journal of
  Photogrammetry and Remote Sensing}, {\bf 191},  143--154, (2022).

\bibitem{zhang2019few}
Hongguang Zhang, Jing Zhang, and Piotr Koniusz, `Few-shot learning via
  saliency-guided hallucination of samples', in {\em Proceedings of the
  IEEE/CVF Conference on Computer Vision and Pattern Recognition}, pp.
  2770--2779, (2019).

\end{thebibliography}
\end{document}